\documentclass[10pt,twocolumn,letterpaper]{article}

\usepackage{cvpr}              

\usepackage{graphicx}
\usepackage{amsmath}
\usepackage{amssymb}
\usepackage{booktabs}

\usepackage{bm}
\usepackage{algorithm}
\usepackage{algorithmic}
\usepackage{multirow}
\usepackage{arydshln}
\usepackage{threeparttable}
\usepackage[accsupp]{axessibility}

\usepackage[pagebackref,breaklinks,colorlinks]{hyperref}

\usepackage[capitalize]{cleveref}
\crefname{section}{Sec.}{Secs.}
\Crefname{section}{Section}{Sections}
\Crefname{table}{Table}{Tables}
\crefname{table}{Tab.}{Tabs.}

\def\cvprPaperID{2018} 
\def\confName{CVPR}
\def\confYear{2022}

\begin{document}


\title{Deep Anomaly Discovery from Unlabeled Videos via\\ Normality Advantage and Self-Paced Refinement}


\author{
Guang Yu\footnotemark[1],
~~
Siqi Wang\footnotemark[1]~\footnotemark[2], 
~~
Zhiping Cai\footnotemark[2], 
~~
Xinwang Liu,
~~
Chuanfu Xu,
~~
Chengkun Wu
\\
National University of Defense Technology, China\\
\tt\small \texttt{\{guangyu, wangsiqi10c, zpcai, xinwangliu,  xuchuanfu, chengkun\_wu\}@nudt.edu.cn}
}
\maketitle
\renewcommand{\thefootnote}{\fnsymbol{footnote}}
\footnotetext[1]{Equal contribution.}
\footnotetext[2]{Corresponding author.}

\begin{abstract}
   While classic video anomaly detection (VAD) requires labeled normal videos for training, emerging unsupervised VAD (UVAD) aims to discover anomalies directly from fully unlabeled videos. However, existing UVAD methods still rely on shallow models to perform detection or initialization, and they are evidently inferior to classic VAD methods. This paper proposes a full deep neural network (DNN) based solution that can realize highly effective UVAD. First, we, for the first time, point out that deep reconstruction can be surprisingly effective for UVAD, which inspires us to unveil a property named ``normality advantage’’, i.e., normal events will enjoy lower reconstruction loss when DNN learns to reconstruct unlabeled videos. With this property, we propose Localization based Reconstruction (LBR) as a strong UVAD baseline and a solid foundation of our solution. Second, we propose a novel self-paced refinement (SPR) scheme, which is synthesized into LBR to conduct UVAD. Unlike ordinary self-paced learning that injects more samples in an easy-to-hard manner, the proposed SPR scheme gradually drops samples so that suspicious anomalies can be removed from the learning process. In this way, SPR consolidates normality advantage and enables better UVAD in a more proactive way. Finally, we further design a variant solution that explicitly takes the motion cues into account. The solution evidently enhances the UVAD performance, and it sometimes even surpasses the best classic VAD methods. Experiments show that our solution not only significantly outperforms existing UVAD methods by a wide margin (5\% to 9\% AUROC), but also enables UVAD to catch up with the mainstream performance of classic VAD.

\end{abstract}

\begin{figure}
	\centering
	\begin{subfigure}[c]{1\linewidth}
		\centering
		\includegraphics[scale=0.95]{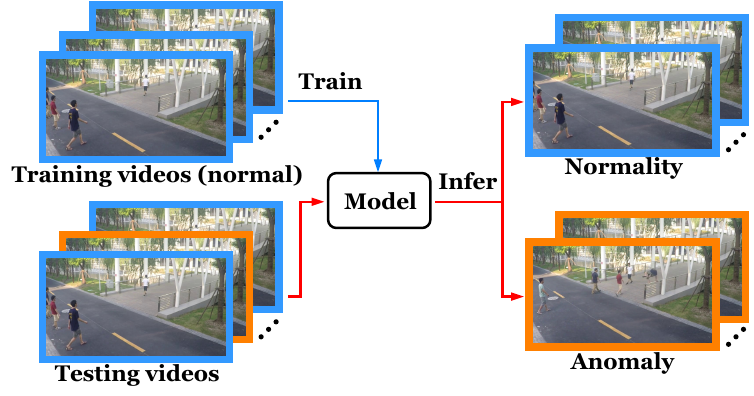}
		\caption{Classic video anomaly detection.}
		\label{fig:classic_vad}
	\end{subfigure}
	
	\begin{subfigure}[c]{1.03\linewidth}
		\centering
		\includegraphics[scale=0.95]{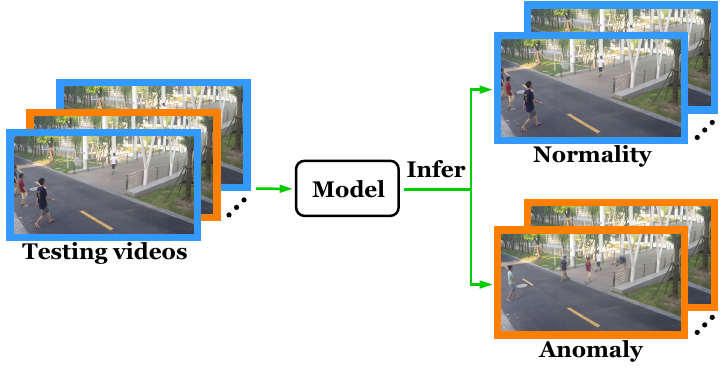}
		\caption{Unsupervised video anomaly detection (UVAD).}
		\label{fig:uvad}
	\end{subfigure}
	\caption{Comparison of classic VAD and UVAD.}
\end{figure}

\section{Introduction}
\label{sec:intro}
Video anomaly detection (VAD) \cite{ramachandra2020survey,jimaging4020036} has constantly been a valuable topic in computer vision, as it aims to automatically discover abnormal events (i.e., anomalies) that deviate from frequently-seen normal routine in surveillance videos. With a great potential to be applied to realms like public security and city management \cite{Pang2020SelfTrainedDO,yu2020cloze}, VAD enjoys a continuous interest from both academia and industry. However, VAD remains open and unsolved. The underlying reason is that anomalies are typically rare and novel, and such characteristics make anomalies hard to be foreseen or enumerated in practice. As a result, a sufficient and comprehensive collection of anomaly data can be particularly difficult or even impossible, which makes the fully supervised classification paradigm not directly applicable to VAD. 

Thus, \textbf{\textit{classic VAD}} follows a \textit{semi-supervised} setup, which labels a training set that contains only normal videos to train a normality model (see Fig. \ref{fig:classic_vad}). During inference, video events that do not fit this normality model are viewed as anomalies. Although such a classic semi-supervised VAD paradigm avoids the thorny issue to collect anomaly data, it still requires human efforts to label a training set with pure normal events. The labeling process can also be particularly tedious and labor-intensive, especially when faced with surging surveillance videos. To alleviate this problem, a natural idea is to perform \textbf{\textit{unsupervised VAD}} (UVAD), which aims to discover anomalies directly from fully unlabeled videos in an unsupervised manner (see Fig. \ref{fig:uvad}). In this way, UVAD no longer requires labeling normal videos to build a training set, which can significantly reduce the cost of time and labor. Therefore, several recent works \cite{del2016discriminative,tudor2017unmasking,liu2018classifier,Pang2020SelfTrainedDO} have explored this topic as a promising alternative to classic VAD (reviewed in Sec. \ref{sec:related_work_VAD}).


Despite some progress, we notice that existing UVAD solutions suffer from two prominent limitations: \textbf{(1)} \textit{Existing UVAD methods typically rely on shallow models to perform detection or initialization, and most of them still involve hand-crafted feature descriptors.} To be more specific, the core idea of representative UVAD methods \cite{del2016discriminative,tudor2017unmasking,liu2018classifier} is to detect drastic changes as anomalies, which often involves learning a shallow detection model (e.g., logistic regression) with descriptor (e.g., 3D gradients) based video representations. However, the expressive power of both the shallow model and hand-crafted descriptors can be limited. The latest work \cite{Pang2020SelfTrainedDO} for the first time introduces deep neural networks (DNNs) to avoid hand-crafted descriptors, but it must resort to an initialization step that involves an isolation forest \cite{liu2008isolation} model to obtain initial results. \textbf{(2)} \textit{The performance of existing UVAD methods is evidently inferior to classic VAD methods.} Taking the commonly-used UCSDped1 and UCSDped2 dataset for an example, recent classic VAD methods usually lead existing UVAD methods by about 10\% AUROC. Meanwhile, existing UVAD methods typically report their performance on earlier datasets, while their applicability and effectiveness on recent benchmark dataset like ShanghaiTech \cite{liu2018future} are also unknown.


To move beyond the above limitations, we propose a novel DNN based solution that can perform UVAD in a highly effective and fully end-to-end manner. Specifically, this paper contributes to UVAD in terms of three aspects:

\begin{itemize}
	\item We, for the first time, point out that deep reconstruction is actually surprisingly effective for UVAD, while such effectiveness further motivates us to unveil the property named \textit{``normality advantage’’}. Based on such a property, we design Localization based Reconstruction (LBR), which serves as a strong deep UVAD baseline and the solid foundation of our deep UVAD solution. 
	\item We design a novel self-paced refinement (SPR) scheme, which is synthesized into LBR to consolidate normality advantage and enable more proactive UVAD. Unlike ordinary self-paced learning (SPL) that gradually injects training samples from easy to hard, the proposed SPR scheme aims to drop suspicious samples, so as to remove anomalies and focus on learning with normality. To our best knowledge, this is also the first attempt to tailor SPL for addressing VAD.
	\item  We further design a motion enhanced solution that explicitly takes the motion cues into account. The variant solution can consistently enhance the detection capability, and sometimes even allows our UVAD solution to outperform state-of-the-art classic VAD methods. 
\end{itemize}

Experiments demonstrate the remarkable advantage of our solution against its UVAD counterparts. Furthermore, it for the first time achieves readily comparable performance to recent classic VAD methods on mainstream benchmarks. 



\begin{figure*}
	\centering
	\begin{subfigure}[c]{0.48\linewidth}
		\centering
		\includegraphics[scale=0.6]{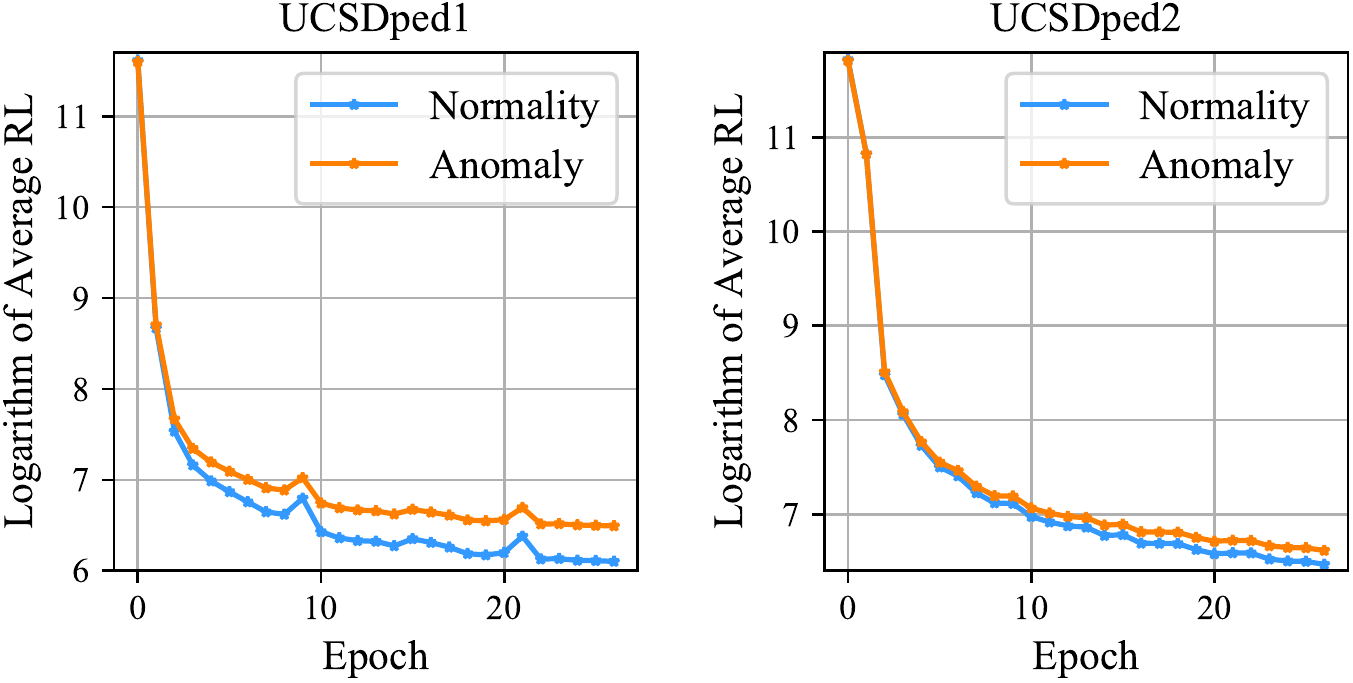}
		\caption{Average reconstruction loss (RL) of normal and abnormal frames.}
		\label{fig:frame_epoch_loss}
	\end{subfigure}
	\hspace{1mm}
	\begin{subfigure}[c]{0.48\linewidth}
		\centering
		\includegraphics[scale=0.6]{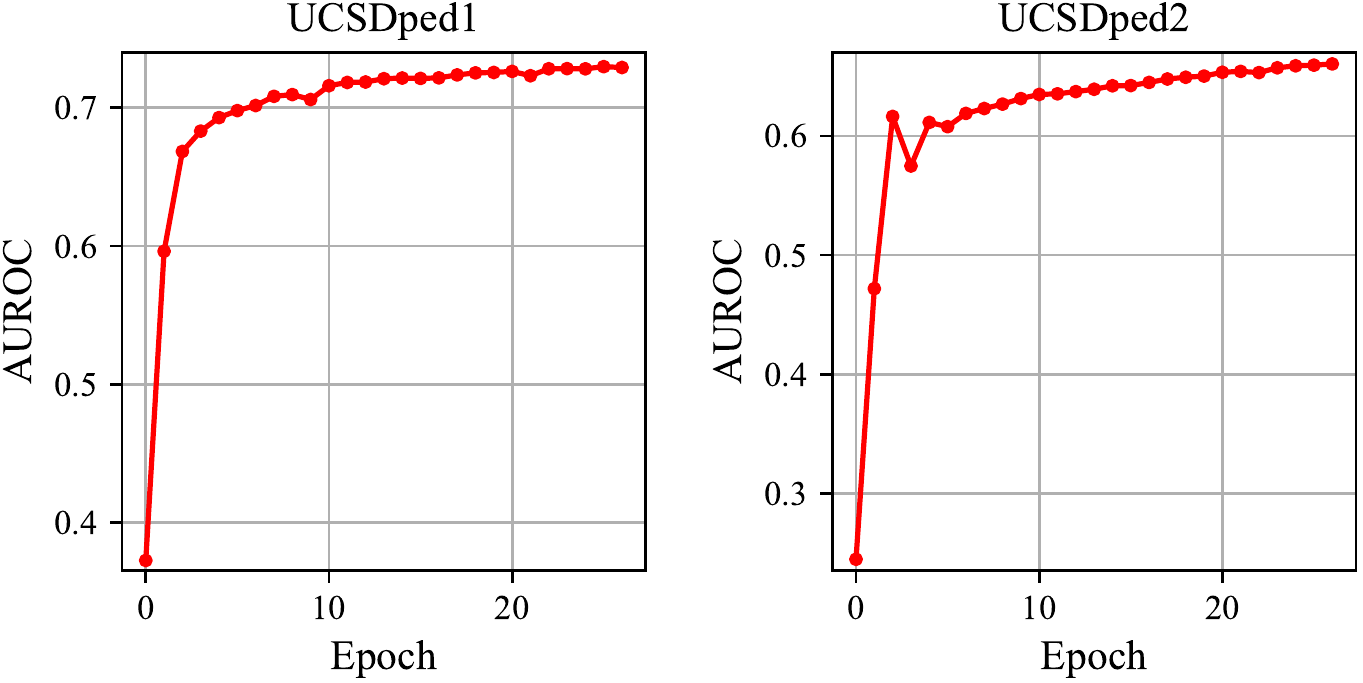}
		\caption{Frame-level Area Under ROC Curve (AUROC) during training.}
		\label{fig:frame_epoch_auc}
	\end{subfigure}
	\caption{A demonstration of normality advantage by FBR on the testing set of UCSDped1 and UCSDped2 dataset.}
\end{figure*}
\section{Related Work}
\label{sec:related_work}
\subsection{Video Anomaly Detection (VAD)}
\label{sec:related_work_VAD}


\textbf{Classic VAD}. Early classic VAD methods usually consist of two steps: First, they utilize hand-crafted feature descriptors (e.g., trajectory \cite{piciarelli2008trajectory}, dynamic texture \cite{mahadevan2010anomaly}, histogram of optical flow \cite{cong2011sparse}, 3D gradients \cite{lu2013abnormal}) to represent original training videos. Then, the extracted features are fed into a shallow normality model for training and inference, such as sparse reconstruction models \cite{cong2011sparse,lu2013abnormal,zhao2011online}, probabilistic models \cite{mahadevan2010anomaly,cheng2015video}, one-class classifiers \cite{2017Video} and nature inspired models \cite{Mehran2009Abnormal,2016Online}. As manual descriptor design can be troublesome and inflexible, recent works are naturally motivated to introduce DNNs for automatic representation learning and end-to-end VAD. Thus, DNN based classic VAD methods have enjoyed a surging interest and explosive development \cite{10.1007/978-3-319-68548-9_70,ravanbakhsh2017abnormal,Ravanbakhsh2018PlugandPlayCF,Ionescu2019DetectingAE,morais2019learning,Lai2020VideoAD,Ramachandra2020LearningAD,Markovitz2020GraphEP,Yu2021AbnormalED,Ramachandra2021PerceptualML}. Due to the absence of anomalies in training, they usually build a DNN normality model by training the DNN to perform some surrogate learning tasks, such as reconstruction \cite{xu2017detecting,tran2017anomaly,yan2018abnormal,Wu2020FastSC} and prediction \cite{lu2019future,Rodrigues2020MultitimescaleTP,Zhang2020NormalityLI,Chen2020AnomalyDI,Li2020VideoFP,DOSHI2021107865}. To improve representation learning and normality modeling, various DNN models have been explored such as recurrent neural networks \cite{luo2017revisit,Luo2021VideoAD} and generative adversarial network \cite{ravanbakhsh2017abnormal,sabokrou2018adversarially,Zaheer2020OldIG}. A more detailed review on classic VAD can be found in \cite{ramachandra2020survey}. Besides, note that deep VAD in this paper refers to directly learning from pixel-level video data by DNNs for VAD.


\textbf{UVAD}. Compared with thoroughly-studied classic VAD, only limited works have explored this emerging topic: Del et al. \cite{del2016discriminative} pioneer the exploration of UVAD by detecting drastic changes as anomalies. Specifically, they describe each video frame by hand-crafted descriptors, and then train a shallow classifier to differentiate two temporally consecutive set of features. Afterwards, an easy classification indicates a drastic change, while shuffling is used to make the classification order-independent; Ionescu et al. \cite{tudor2017unmasking} follow the direction of \cite{del2016discriminative}, but improve change detection by a more sophisticated unmasking scheme: With features calculated by hand-crafted descriptor and pre-trained DNN, they iteratively remove the most discriminative feature in classification. Frames that are still easy to classify after several rounds of removal are viewed as anomalies; Liu et al. \cite{liu2018classifier} study the connection between unmasking and statistical learning, and further enhance the performance by a history sampling method and a new frame-level motion feature. Unlike above methods that are essentially based on the change detection paradigm, the latest work from Pang et al. \cite{Pang2020SelfTrainedDO} first obtain the preliminary detection results by leveraging a pre-trained DNN and an isolation forest \cite{liu2008isolation}. Results are then refined by performing a two-class ordinal regression in a self-trained fashion. 



\textbf{Weakly-supervised VAD (WVAD)}. WVAD has been another heated topic \cite{Sultani2018RealWorldAD,Zhong_2019_CVPR,Liu2019MarginLE,Zaheer2020CLAWSCA,Feng_2021_CVPR,Tian_2021_ICCV,Purwanto_2021_ICCV,Lee_2021_ICCV,Wu2021WeaklySupervisedSA} in current research. Unlike classic VAD or UVAD, WVAD utilizes video-level annotations for training, so as to reduce the cost of labeling \cite{Sultani2018RealWorldAD}. Since WVAD usually adopts a different setup and benchmarks from most classic VAD and UVAD works, we will not discuss WVAD in this paper.



\subsection{Self-Paced Learning}
\label{subsec:related_work_SPL}
Self-paced learning (SPL) is a branch of curriculum learning (CL) \cite{Wang2021ASO,Soviany2021CurriculumLA}. Motivated by the beneficial learning order in human curricula, CL introduces a learning strategy that trains the model with samples in an easy-to-hard manner \cite{Bengio2009CurriculumL}. To avoid the manual design of difficulty measures in classic CL, SPL is proposed to automatically measure the difficulty of samples based on the training losses \cite{Kumar2010SelfPacedLF}. Specifically, given a leaning objective, SPL embeds learnable sample weights and a self-paced (SP) regularizer into the objective. The SP regularizer enables SPL to learn a proper weight for each sample, so as to control the curriculum of learning. As a center issue of SPL, the design of SP regularizer has been extensively studied \cite{Kumar2010SelfPacedLF,Jiang2014EasySF,Jiang2014SelfPacedLW,Zhao2015SelfPacedLF,Xu2015MultiviewSL,Li2017SelfpacedCN,Fan2017SelfPacedLA,Gong2019DecompositionBasedEM}, and the plug-and-play nature of SPL enables it to be widely applied to various tasks, such as classification \cite{Tang2012SelfpacedDL,10.1007/978-3-030-58558-7_15}, object segmentation \cite{Zhang2017SPFTNAS}, domain adaptation\cite{Zhang2021SelfPacedCA}, object detection \cite{Zhang2018LeveragingPF,Sangineto2019SelfPD}, clustering \cite{Ghasedi2019BalancedSL,Guo2020AdaptiveSD}, object re-identification \cite{NEURIPS2020_821fa74b}. However, to our best knowledge, none of existing works has explored SPL for VAD.

\section{The Proposed UVAD Solution}
\label{sec:method}

\subsection{Reconstruction in Classic VAD}
\label{sec:ND}

Although our goal is to develop a deep UVAD solution, it will be helpful to recall how DNN addresses classic VAD in the first place. Owing to the lack of anomalies in training, DNN cannot learn representations directly by supervised classification. Instead, reconstruction has been a frequently-used deep learning paradigm for classic VAD. Typically, the reconstruction paradigm learns to embed the normal training video $\mathbf{x}$ into a low-dimensional embedding by an encoder network $f_e(\cdot)$, and then reconstruct the input video from the embedding by a decoder network $f_d(\cdot)$. This goal is often realized by solving the objective below:


\begin{equation}
    \label{eq:recon}
    \min_{\bm{\theta}}\sum_{\mathbf{x}} L_R(f_d(f_e(\mathbf{x})), \mathbf{x}|\bm{\theta})+R(\bm{\theta})
\end{equation}
where $\bm{\theta}$ denotes all learnable parameters of the encoder and decoder, and $L_R(\cdot, \cdot|\bm{\theta})$ is a loss function that measures the reconstruction loss (RL) under parameters $\bm{\theta}$. $R(\bm{\theta})$ is a regularization term that prevents overfitting. By Eq. (\ref{eq:recon}), DNN is expected to learn normality patterns and reconstruct normal events well, while large RL is produced for unseen anomalies. As a straightforward deep learning paradigm, reconstruction is extensively applied to classic VAD \cite{ramachandra2020survey}. 

\subsection{Normality Advantage in UVAD}
\label{sec:na}
Despite the popularity of DNN based reconstruction in classic VAD, it has not been explored as a deep solution to UVAD. Seemingly, learning by unlabeled videos mixed with anomalies also enables DNN to reconstruct anomalies, which disables it from discriminating anomalies. However, we argue that it may not be true: {In most cases, anomalies are unusual events that occur at a low probability, while the majority of events in videos are still normal}. When DNN learns to reconstruct unlabeled videos that contain anomalies, \textit{the imbalanced nature of normality/anomaly tends to bias the DNN model towards the majority class (normality), which offers us a chance to differentiate normality and anomalies}. Besides, we also notice that such bias is reported in simulated outlier image removal experiments \cite{Xia2015LearningDR,Wang2019EffectiveEU}. 


\begin{figure}
	\centering
	\includegraphics[scale=0.9]{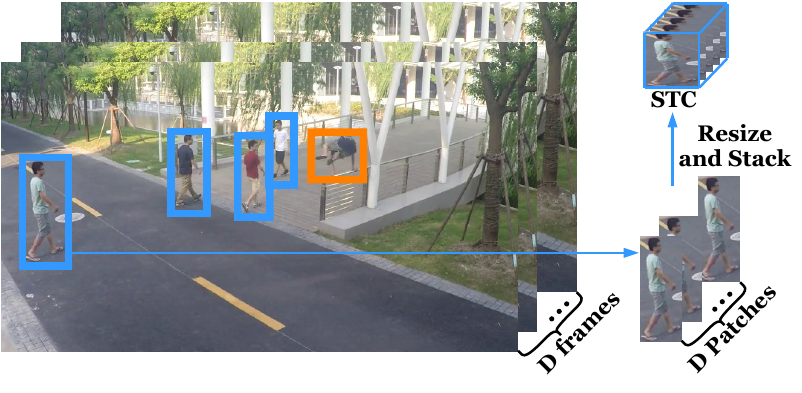}
	\caption{Localizing foreground to build spatio-temporal cube.}
	\label{fig:localization}
\end{figure}

Motivated by such an intuition, we conduct some basic experiments to test whether DNN based reconstruction can be a feasible deep solution to UVAD: Following most UVAD works \cite{del2016discriminative,tudor2017unmasking,liu2018classifier}, we directly use the testing set of a VAD benchmark dataset as unlabeled videos with anomalies, while both the training set and testing set labels are strictly unused when training the DNN. To perform reconstruction, we train a multi-layer fully convolutional autoencoder (CAE) network to reconstruct the frames of unlabeled videos. To evaluate the reconstruction of normal and abnormal frames, we compute the average RL of normal frames and abnormal frames respectively. As an example, we visualize the logarithm of the average RL on UCSDped1 and UCSDped2 dataset in Fig. \ref{fig:frame_epoch_loss}, and some interesting observations can be drawn: Initially, the averaged RL of normal and abnormal events are very close. Afterwards, a loss gap gradually appears between normal and abnormal frames, which suggests that DNN prioritizes the reconstruction of normality. Moreover, the gap persists to exist as the training continues. Such observations lead to an interesting conclusion: \textbf{\textit{Normality tends to play a more advantageous role (i.e., enjoys a lower reconstruction loss) when DNN learns to reconstruct both normality and anomalies in unlabeled videos}}, which is named as \textbf{\textit{normality advantage}} of UVAD.



To further validate whether normality advantage can be utilized to discriminate anomalies, we simply use RL as the anomaly score of each video frame, and calculate frame-level AUROC \cite{mahadevan2010anomaly} to quantitatively evaluate the VAD performance during the learning process: As shown in Fig. \ref{fig:frame_epoch_auc}, whilst the VAD performance is poor at the beginning, it will be rapidly improved in 3-5 starting epochs. Afterwards, the AUROC tends to increase slowly and gradually levels off. As a consequence, those observations demonstrate the possibility to exploit normality advantage for deep UVAD. In addition, we would like to make the following remarks: \textbf{(1)} Normality advantage stems from the dominant role of normal events in videos. This role is essentially decided by the nature of anomalies, which are supposed to be rare events that divert from the majority. Actually, when a certain anomaly becomes frequent, they should be viewed as the new normality. Thus, we simply assume that normality advantage usually holds in the context of UVAD. \textbf{(2)} In Sec. \ref{sec:discussion}, we will show that other deep learning paradigms (e.g., prediction) can also exploit this property to perform UVAD. This paper will focus on reconstruction as it is one of the most frequently-used deep paradigms in VAD. 



\subsection{Localization based Reconstruction (LBR)}
\label{sec:LBR}
\begin{figure}
	\centering
	\includegraphics[scale=0.59]{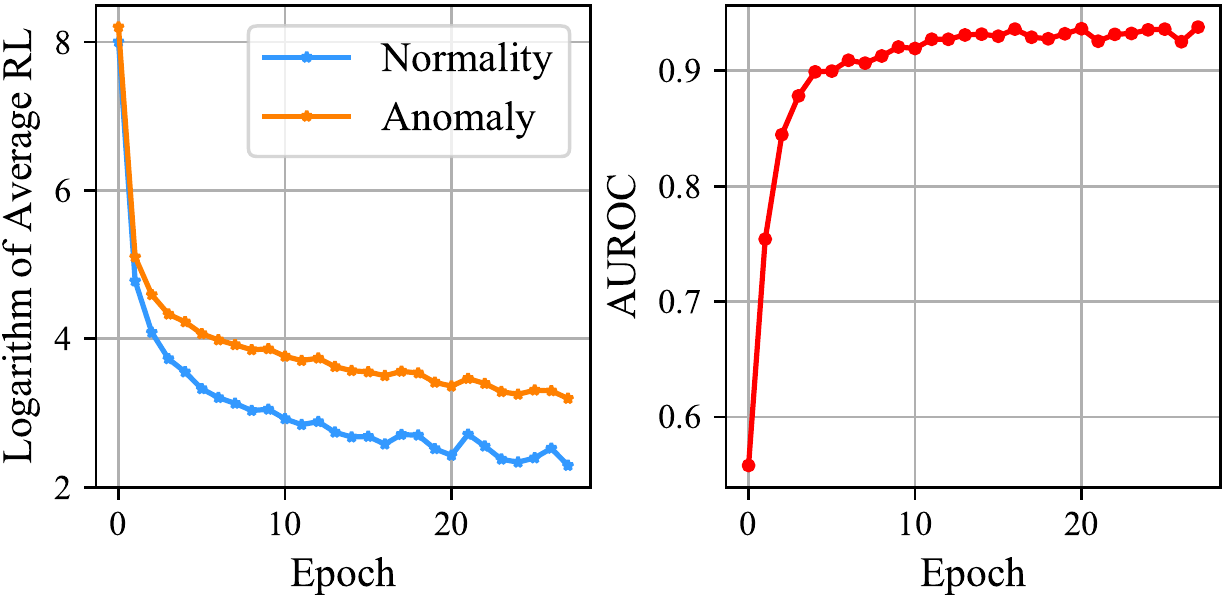}
	\caption{Average RL of normal/abnormal STCs (left) and frame-level AUROC (right) of LBR on UCSDped2 dataset.}
	\label{fig:ped2_FBR_epoch_results}
\end{figure}

Normality advantage renders frame based reconstruction (FBR) a feasible deep solution to UVAD, but its performance is still inferior to existing UVAD methods. For example, the RL gap of FBR on UCSDped2 dataset is relatively small (see Fig. \ref{fig:frame_epoch_loss}), while its AUROC is also unsatisfactory. Actually, there is an important reason for its unsatisfactory performance: {In many cases, only a small region of a video frame is anomalous, while the remaining part is still normal.} Thus, FBR is obviously not the optimal way to manifest normality advantage, since video events cannot be precisely represented on a per-frame basis. Inspired by recent works \cite{hinami2017joint,ionescu2019object,yu2020cloze} that explore localization for classic VAD, we propose to introduce localization as a remedy to the drawback of FBR. Although localization is first introduced by classic VAD, we must point out that localization brings one unique benefit to UVAD: \textit{Localization is able to magnify the normality advantage when performing UVAD}. An example is shown in Fig. \ref{fig:localization}: Consider a video frame with four walking pedestrians (normality) and one fence jumper (anomaly). For frame based analysis, the entire frame will be viewed as one abnormal event. By contrast, localization enables us to extract four normal events and one abnormal event. In this way, more normal events will exhibit a larger advantage against the anomalies in reconstruction.  


Following this idea, we propose \textit{localization based Reconstruction} (LBR) as a new deep baseline for UVAD: As to localization, we follow the localization scheme proposed in \cite{yu2020cloze}, which is shown to achieve both precise and comprehensive localization (the procedure is detailed in supplementary material). For each localized object on the frame, we extract $D$ patches from the current and adjacent $(D-1)$ frames. Extracted patches are resized into $H\times W$, and then stacked into a $H\times W\times D$ spatio-temporal cube (STC), which is used to represent a video event (illustrated in Fig. \ref{fig:localization}). The DNN is then trained to reconstruct extracted STCs, while RL of STCs are also used as anomaly scores. To perform frame-level evaluation, the maximum of all STCs' scores on a frame is considered as the score of this frame. To illustrate how LBR magnifies normality advantage, we visualize LBR's average RL of normality/anomaly and AUROC in training on UCSDped2 dataset, on which FBR performs poorly. As shown in Fig. \ref{fig:ped2_FBR_epoch_results}, LBR enjoys a remarkably larger RL gap than FBR, while the frame-level AUROC also grows to over $90\%$. In Fig. \ref{fig:frame_compare_auc},  we further compare frame-level AUROC of FBR and LBR (detailed in Sec. \ref{sec:exp_setting}) with existing UVAD methods on several commonly-used VAD benchmarks, and find that LBR is surprisingly effective: As a straightforward baseline, LBR has already been able to outperform all existing UVAD methods on those benchmarks. Meanwhile, LBR achieves a large performance gain when compared with FBR, which verifies the importance of localization for UVAD. Consequently, the proposed LBR is able to lay a solid foundation for our deep UVAD solution.

\subsection{Self-Paced Refinement (SPR)}

Although LBR is shown to be a strong UVAD baseline, it passively relies on normality advantage to detect anomalies, and anomalies are constantly reserved in training. Nevertheless, the proactive removal of anomalies is obviously more preferable. To be more specific, we intend to sort out suspicious anomalies by RL and actively reduce anomalies' influence on DNN, so as to refine the DNN model and consolidate the normality advantage. To this end, we notice that self-paced learning (SPL) \cite{Kumar2010SelfPacedLF} provides an elegant strategy to adjust the influence of each individual sample in learning. However, traditional SPL usually injects harder samples to training in an incremental manner, but our goal is to gradually remove suspicious anomalies from the given data. To bridge this gap, we design a novel \textit{Self-Paced Refinement} (SPR) scheme for UVAD, which is detailed below:

\begin{figure}[t]
	\centering
	\includegraphics[scale=0.48]{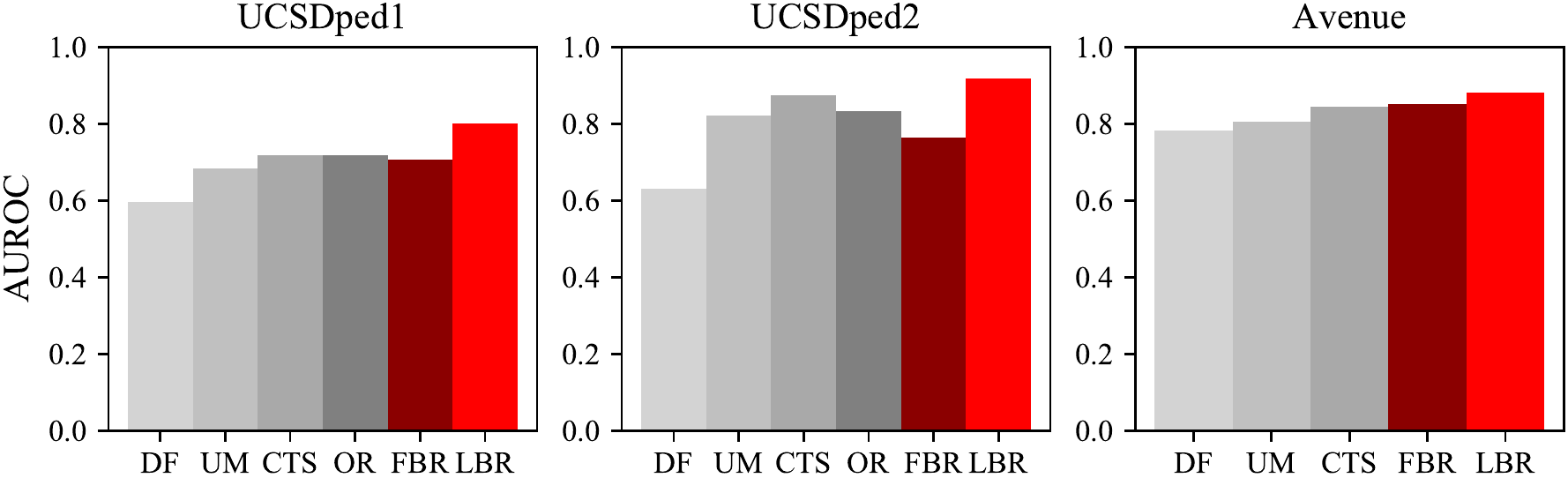}
	\caption{AUROC comparison between FBR/LBR and existing UVAD methods (DF \cite{del2016discriminative}, UM \cite{tudor2017unmasking}, CTS \cite{liu2018classifier}, OR \cite{Pang2020SelfTrainedDO}).}
	\label{fig:frame_compare_auc}
\end{figure}
\label{sec:SPR}



We first review the ordinary SPL as preliminaries. Formally, let $\mathcal{D}=\{(\mathbf{x}_i,y_i)\}_{i=1}^{N}$ denote the training set, where $\mathbf{x}_i$ and $y_i$ represent $i$-th sample and its learning target respectively. A model $f$ parameterized by $\bm{\theta}$ maps a sample $\mathbf{x}_i$ to a prediction $f(\mathbf{x}_i)$, while the training loss $L(f(\mathbf{x}_i), y_i)$ is calculated by some loss function $L$. The learning goal is usually written as the following objective:
 \begin{equation}
 \min\limits_{\bm{\theta}}\sum_{i=1}^{N}L(f(\mathbf{x}_i), y_i|\bm{\theta})
 \label{eq:origin_objective}
 \end{equation}
Note that we omit the regularization term $R$ for simplicity. As to SPL, it embeds the learnable sample weights $\mathbf{v}=[v_i,\dots,v_N]\in[0,1]^N$ and a self-paced (SP) regularizer $g(\mathbf{v}|\lambda)$ into the above learning objective, where $\lambda$ is an age parameter to control the learning pace. Specifically, the goal of SPL is to solve the optimization problem below:
 \begin{equation}
\min\limits_{\bm{\theta},\mathbf{v}} \sum_{i=1}^{N}v_i L(f(\mathbf{x}_i),y_i|\bm{\theta})+g(\mathbf{v}|\lambda)
 \label{eq:spl_objective}
 \end{equation}
Eq. (\ref{eq:spl_objective}) can be solved by an alternative search strategy (ASS) \cite{Kumar2010SelfPacedLF}, which alternatively optimizes $\bm{\theta}$ or $\mathbf{v}$ while keeping the other fixed. To facilitate optimization of $\mathbf{v}$, the SP regularizer $g(\mathbf{v}|\lambda)$ is usually designed to be convex, so when fixing $\bm{\theta}$ the global minimum $\mathbf{v}^*$ can be easily yielded by setting the partial derivative to be 0. It can be shown that $\mathbf{v}^*$ is usually determined by the training loss $L(f(\mathbf{x}_i),y_i)$ and age parameter $\lambda$. To enable SPL, $\lambda$ is usually initialized by a small value, which produces $\mathbf{v}^*$ that only involves a few easy samples with small loss at the early training stage. Then, $\lambda$ is gradually increased to introduce harder samples into training until all samples are considered in the end.

As shown above, SPL can adjust the weights of samples by considering their hardness and the current learning stage. Such desirable abilities make SPL perfectly eligible for enlarging normality advantage, which can be realized by assigning smaller weights to suspicious anomalies with large RL. Thus, we develop SPR from SPL: Concretely, given a sampled batch of STCs $\{\mathbf{c}_i\}_{i=1}^{n}$ ($\mathbf{c}_i$ denotes the $i$-th STC), SPR minimizes an objective $\mathcal{L}_{SPR}$ w.r.t. the DNN parameters $\bm{\theta}$ and sample weights $\mathbf{v}$, while $\mathcal{L}_{SPR}$ is defined by:

\begin{equation}
\label{eq:spr}
\mathcal{L}_{SPR} = \sum_{i=1}^{n}v_i L_i(\bm{\theta}) + g(\mathbf{v}|\lambda)
\end{equation}
where $L_i(\bm{\theta})=L_R(f_d(f_e(\mathbf{c}_i)), \mathbf{c}_i|\bm{\theta})$ represents the RL of $\mathbf{c}_i$, and the regularization term $R(\bm{\theta})$ is also omitted for simplicity. As mentioned above, Eq. (\ref{eq:spr}) is optimized by ASS: When $\mathbf{v}$ is fixed, the objective can be transformed into: 
\begin{equation}
\min_{\bm{\theta}}\sum_{i=1}^{n}v_i L_i(\bm{\theta})
\label{eq:opt_theta}
\end{equation}
The goal in Eq. (\ref{eq:opt_theta}) can be optimized by gradient descent. In fact, it assigns a weight to each STC when DNN learns to reconstruct STCs, which encourages DNN to place more emphasis on reconstructing the STC with larger weight $v_i$. When $\bm{\theta}$ is fixed, the optimal ${v}_i^*$ can be obtained by solving:
\begin{equation}
\min_{v_i\in[0,1]}\sum^n_{i=1} v_i L_i(\bm{\theta})+ g(\mathbf{v}|\lambda)
\label{eq:opt_v}
\end{equation}
Qualitatively, our SPR expects the optimal sample weight $v^*_i$ yielded by Eq. (\ref{eq:opt_v}) to meet the following requirements: When the loss of a STC is very large/small among its peers, it is highly likely to be abnormal/normal. Accordingly, its sample weight $v_i$ should be directly set to 0/1. Otherwise, the sample weight should be negatively correlated with its likelihood to be abnormal, which is embodied by its RL. Such requirements motivate us to leverage a mixture SP regularizer \cite{Jiang2014EasySF} for SPR, which is in the following form:

\begin{equation}
g(\mathbf{v}|\lambda,\lambda') = -\rho \sum_{i=1}^{n}\ln(v_i + \frac{\rho}{\lambda})
\label{eq:mixture}
\end{equation}
where $\lambda'$ is an additional parameter that satifies $\lambda>\lambda'>0$, and $\rho=\frac{\lambda\lambda'}{\lambda-\lambda'}$. As the mixture SP regularizer is convex, $v_i^*$ to Eq. (\ref{eq:opt_v}) can be derived by setting the partial derivative of $\mathcal{L}_{SPR} $ w.r.t $v_i$ to zero, which yields:


\begin{equation}
\frac{\partial \mathcal{L}_{SPR}}{\partial v_i} = L_i(\bm{\theta}) - \frac{\rho}{v_i+\frac{\rho}{\lambda}} = 0,\quad i=1,\cdots,n
\label{eq:derivative}
\end{equation}
Based on Eq. (\ref{eq:derivative}) and the constraint $v_i \in [0,1]$, a closed-formed solution to Eq. (\ref{eq:opt_v}) can be derived as follows:
\begin{equation}
\label{eq:v_solution}
v_i^* = \left\{
\begin{aligned} 
&0, \quad L_i(\bm{\theta}) \geq \lambda\\
&\frac{\rho}{L_i(\bm{\theta})} - \frac{\lambda'}{\lambda-\lambda'}, \quad \lambda'<L_i(\bm{\theta})<\lambda\\
&1, \quad L_i(\bm{\theta}) \leq \lambda'
\end{aligned}
\right.
\end{equation}
From Eq. (\ref{eq:v_solution}), we can see how SPR consolidates normality advantage and excludes anomalies in an active way: When the RL of a STC $L_i(\bm{\theta})$ is larger than a upper threshold $\lambda$, its weight $v_i$ will be directly set to 0, which suggests that this STC will be directly dropped from the current iteration. Similarly, the weight of STC will be directly set to 1 when its RL is smaller than a lower threshold $\lambda'$, which enables it to fully participate in learning. For those STCs that are less certain ($\lambda'<L_i(\bm{\theta})<\lambda$), their weights are inversely proportional to their RL. Next, the most important issue is to determine $\lambda$ and $\lambda'$, and we propose a self-adaptive strategy to calculate them by the statistics of RL: At the $t$-th iteration of model updating, the lower threshold $\lambda'=\mu(t)+\sigma(t)$, where $\mu(t)$ and $\sigma(t)$ denote the mean and standard deviation of STCs' RL in the current batch. The design of $\lambda'(t)$ indicates that we expect the majority of events to be normal. As to the upper threshold $\lambda$, we set it as follows:


\begin{equation}
\lambda = \max\{\mu(t) + (4 - t \cdot r)\cdot \sigma(t), \lambda'\}
\label{eq:lambda}
\end{equation}
where $r$ is the shrink rate that usually takes a small value. The intuition behind $\lambda$ is also straightforward: At the beginning, we only view STCs with very high RL ($L_i(\bm{\theta})\geq\mu(t)+4\sigma(t)$) as certain anomalies. As the learning continues, the normality advantage becomes more evident and allows us to exclude more anomalies. Thus, as $t$ increases, Eq. (\ref{eq:lambda}) enables us to gradually shrink the coefficient of $\sigma(t)$ until $\lambda$ decreases to $\lambda'$, so as to exclude a larger portion of suspicious anomalies. Since the initial RL is not informative, SPR is introduced after a few warm-up epochs, which allows normality to establish the preliminary advantage. The whole SPR scheme is presented in Algorithm \ref{alg:SPR}.

\begin{algorithm}[t]
    \footnotesize
	\caption{Self-Paced Refinement}
	\label{alg:SPR}
	\begin{algorithmic}[1] 
		\REQUIRE A DNN $f$ with parameters $\bm{\theta}$, the set $\mathcal{C}$ of $N$ STCs collected from unlabeled videos, batch size $n$, training epoch $T$, warm-up epoch $T'$
		\ENSURE The updated parameters $\bm{\theta}$
		\STATE Initialize $\bm{\theta}$, $t=0$
		\FOR {$i=1 \to T$}
		\FOR {$j=1 \to \lceil \frac{N}{n}\rceil$}    
		\STATE Randomly sample a batch of $n$ data from $\mathcal{C}$
		\IF {$i\leq T'$}
		\STATE Update $\bm{\theta}$ by Eq. (\ref{eq:recon})
		\ELSE 
		\STATE Compute $\lambda'=\mu(t)+\sigma(t)$ and $\lambda$ by Eq. (\ref{eq:lambda})
		\STATE $t = t + 1$
		\STATE Update $\mathbf{v}$ by Eq. (\ref{eq:v_solution})
		\STATE Updata $\bm{\theta}$ by Eq. (\ref{eq:opt_theta})
		\ENDIF
		\ENDFOR
		\ENDFOR
	\end{algorithmic}
\end{algorithm}

\subsection{Motion Enhanced UVAD Solution}
Many VAD works have pointed out the importance of motion cues \cite{nguyen2019anomaly,yu2020cloze,Liu_2021_ICCV}. Hence, we also design a motion enhanced UVAD solution, which consists of the following steps: First, to represent motion in videos, we adopt the dense optical flow \cite{fortun2015optical}, which depicts pixel-wise motion by estimating the correspondence between two frames. The optical flow map of each video frame can be computed efficiently by a pre-trained DNN model (e.g., FlowNet v2 \cite{ilg2017flownet}). Then, based on the location of each foreground object, we extract $D$ optical flow patches from the optical flow maps that correspond to the current and $(D-1)$ neighboring frames. Similar to the construction of STC, $D$ optical flow patches are resized and stacked into a $H\times W\times D$ optical flow cube (OFC). Then, we introduce a separated motion encoder $f^{(m)}_e$ and decoder $f^{(m)}_d$, which are trained to reconstruct the OFC by taking its corresponding STC as input:


\begin{equation}
    \label{eq:cross_recon}
    \min_{\bm{\theta}'}\sum_{i=1}^n L_R(f^{(m)}_d(f^{(m)}_e(\mathbf{c}_i)), \mathbf{c}_i^{(o)}|\bm{\theta}')+R(\bm{\theta}')
\end{equation}
where $\mathbf{c}_i^{(o)}$ represents the OFC for input STC $\mathbf{c}_i$. $\bm{\theta}'$ is the set of parameters for $f^{(m)}_e$ and $f^{(m)}_d$. After training, the RL of an OFC is computed as the motion anomaly scores $S^{(m)}$. The final anomaly score $S$ is computed as follows:

\begin{equation}
    S(\mathbf{c}_i)=\omega_a \frac{S^{(a)}(\mathbf{c}_i) - \mu^{(a)}}{\sigma^{(a)}} +\omega_m \frac{S^{(m)}(\mathbf{c}_i)-\mu^{(m)}}{\sigma^{(m)}}
\end{equation}
where $S^{(a)}$ is the appearance anomaly score obtained by RL of STCs, and $\mu^{(a)}$, $\sigma^{(a)}$, $\mu^{(m)}$, $\sigma^{(m)}$ are means and standard deviations of appearance/motion anomaly scores for all STCs/OFCs. $\mu^{(a)}$, $\sigma^{(a)}$, $\mu^{(m)}$, $\sigma^{(m)}$ are computable as UVAD handles all testing videos in a transductive manner.

\section{Empirical Evaluations}
\subsection{Experimental Settings}
\label{sec:exp_setting}
We evaluate the proposed UVAD solution (LBR-SPR) on the following commonly-used public VAD datasets: UCSDped1/UCSDped2 \cite{mahadevan2010anomaly}, Avenue \cite{lu2013abnormal} and ShanghaiTech \cite{liu2018future}. To perform UVAD, we adopt two types of UVAD setups in previous UVAD works: \textbf{(1)} \textbf{\textit{Partial mode}} \cite{del2016discriminative,tudor2017unmasking,liu2018classifier}: Only the original testing set of a dataset is used for learning, while the original training set is discarded. \textbf{(2)} \textbf{\textit{Merge mode}} \cite{Pang2020SelfTrainedDO}: The original training set and testing set are merged into one unlabeled set for learning. For both modes, labels are strictly unused in learning. Note that performance evaluation is only conducted on the original testing set of each benchmark, so as to enable comparison with existing VAD methods in the literature. For quantitative evaluation, we adopt the most commonly-used frame-level AUROC \cite{mahadevan2010anomaly} in recent VAD works, while we also introduce and report other metrics like equal error rate (EER) and pixel-level AUROC in supplementary material. To construct STCs and OFCs, we adopt the localization scheme in \cite{yu2020cloze} and set $H=W=32$ and $D=5$. The reconstruction is performed by a 7-layer fully convolutional autoencoder network, which is optimized by the default Adam optimizer in PyTorch toolbox \cite{paszke2019pytorch}. The batch size in training is 256, while RL is computed by mean square error (MSE). For the shrink rate $r$, we adopt $0.0001$ for UCSDped1/Avenue and $0.005$ for UCSDped2/ShanghaiTech. The number of training epochs is set by $T=30$, while $T'=5$ epochs are typically used for warm-up. As to motion enhanced solution, we set $(\omega_a,\omega_m)$ to be $(0.5,1)$ for UCSDped1/UCSDped2/Avenue,  and $(0.1,1)$ for ShanghaiTech. Note that more details are provided in supplementary material due to page limit. 




\begin{table}
	\newcommand{\tabincell}[2]{\begin{tabular}{@{}#1@{}}#2\end{tabular}}
	\centering
	\caption{Frame-level AUROC comparison. Note that LBR-SPR$^*$ indicates the performance of LBR-SPR under partial mode, while LBR-SPR$^+$ indicates the performance under merge mode (explained in Sec. \ref{sec:exp_setting}). ME denotes motion enhancement.}
	\linespread{1}\selectfont
	\scalebox{0.78}{
		\begin{tabular}{|c|c|cccc|}
			\hline
			Setup & Method & Ped1 & Ped2 & Avenue & SHTech \\
			\hline
			\multirow{32}{*}{\rotatebox{90}{Classic VAD}}
			& CAE \cite{hasan2016learning}  & 81.0\%  & 90.0\%  & 70.2\% & - \\
            & ST-CAE \cite{zhao2017spatio}  & 92.3\%  & 91.2\%  & 80.9\% & - \\
			& sRNN \cite{luo2017revisit}  & -   & 92.2\%  & 81.7\% & 68.0\% \\
			& WTA-CAE \cite{tran2017anomaly}  & 91.9\%  & 96.6\%  & 82.1\% & - \\
 			& LSTM-AE \cite{luo2017remembering}  & 75.5\%  & 88.1\%  & 77.0\% & - \\
 			& AM-GAN \cite{ravanbakhsh2017abnormal}  & 97.4\%  & 93.5\% & - & - \\
			& Recounting \cite{hinami2017joint}  & -  & 92.2\% & - & - \\
			& FFP \cite{liu2018future}  & 83.1\%  & 95.4\%  & 85.1\% & 72.8\% \\
			& AnoPCN \cite{ye2019anopcn}  & -  & 96.8\%  & 86.2\% & 73.6\% \\
			& Attention \cite{zhou2019attention}  & 83.9\%  & 96.0\%  & 86.0\% & - \\
			& PDE-AE \cite{abati2019latent}  & -  & 95.4\%  & - & 72.5\% \\
			& Mem-AE \cite{Gong_2019_ICCV}  & -  & 94.1\% & 83.3\% & 71.2\% \\
			& AM-Corr. \cite{nguyen2019anomaly}  & -  & 96.2\%  & 86.9\% & - \\
			& AnomalyNet \cite{zhou2019anomalynet}  & 83.5\%   & 94.9\%  & 86.1\% & - \\
			& Object-Centric \cite{ionescu2019object}  & -  & 97.8\%  & 90.4\% & 84.9\% \\
			& MLAD \cite{Vu2019RobustAD}  & 82.3\%  & 99.2\%  & 71.5\% & - \\
            & BMAN \cite{Lee2020BMANBM} & - & 96.6\% & 90.0\% & 76.2\% \\
			& Clustering-AE \cite{Chang2020ClusteringDD}  & -  & 96.5\% & 86.0\% & 73.3\% \\
			& r-GAN \cite{Lu2020FewshotSA}  & 86.3\%  & 96.2\%  & 85.8\% & 77.9\% \\
 			& DeepOC \cite{Wu2020ADO}  & 83.5\%  & 96.9\%  & 86.6\% & - \\
			& SIGNet \cite{Fang2020AnomalyDW}  & 86.0\%  & 96.2\%  & 86.8\% & - \\
 			& Multipath-Pred. \cite{Wang2021RobustUV}  & 83.4\%   & 96.3\%  & 88.3\% & 76.6\% \\
			& Mem-Guided \cite{Park2020LearningMN}  & -  & 97.0\%  & 88.5\% & 70.5\% \\
 			& CAC \cite{Wang2020ClusterAC}  & - & - & 87.0\% & 79.3\% \\
			& Scene-Aware \cite{Sun2020SceneAwareCR}  & -  & -  & 89.6\% & 74.7\% \\
			& VEC \cite{yu2020cloze}  & -  & 97.3\%  & 90.2\% & 74.8\% \\
			& BAF \cite{Georgescu2021ABF}  & -  & 98.7\%  & 92.3\% & 82.7\% \\
			& AMMCN \cite{Cai2021AppearanceMotionMC}  & -  & 96.6\%  & 86.6\% & 73.7\% \\
			& SSMTL\footnotemark[3] \cite{Georgescu2021AnomalyDI}  & -  & 97.5\%  & 91.5\% & 82.4\% \\
			& MPN \cite{Lv_2021_CVPR}  & 85.1\%  & 96.9\%  & 89.5\% & 73.8\% \\
			& HF$^2$ \cite{Liu_2021_ICCV}  & - & 99.3\%  & 91.1\% & 76.2\% \\
			& CT-D2GAN \cite{Feng2021ConvolutionalTB}  & - & 97.2\%  & 85.9\% & 77.7\% \\
			\hline
			\multirow{8}{*}{\rotatebox{90}{UVAD}} & DF \cite{del2016discriminative}  & 59.6\% & 63.0\% & 78.3\% & - \\
			& UM \cite{tudor2017unmasking} & 68.4\%  & 82.2\%  & 80.6\% & - \\
			& CTS \cite{liu2018classifier}  & 71.8\%  & 87.5\%  & 84.4\% & - \\
			& OR \cite{Pang2020SelfTrainedDO}  & 71.7\%  & 83.2\%  & - & - \\
			\cdashline{2-6}
			& LBR-SPR$^*$ (w/o ME) & \textbf{81.1\%} & 93.3\% & 88.5\% & 71.1\%\\
			& LBR-SPR$^*$ (w/ ME) & \textbf{81.1\%} & \textbf{95.7\%} & \textbf{92.8\%} & \textbf{72.1\%}\\
			\cdashline{2-6}
			& LBR-SPR$^+$ (w/o ME) & 79.4\% & 97.0\% & 89.7\% & 71.9\%\\
			& LBR-SPR$^+$ (w/ ME) & \textbf{80.9\%} & \textbf{97.2\%}  & \textbf{90.7\%} & \textbf{72.6\%}
			\\
			\hline
		\end{tabular}}
	\label{tab:AUC_compare}
\end{table}
\footnotetext[3]{{As micro AUROC is used, we reported results from the official page of \cite{Georgescu2021AnomalyDI} (\url{https://github.com/lilygeorgescu/AED-SSMTL}).}}

\subsection{Comparison with State-of-the-art Methods}

In Table \ref{tab:AUC_compare}, we compare the performance of LBR-SPR with state-of-the-art UVAD solutions. The performance of recent classic VAD methods is also included, while they are listed here as a reference. From Table \ref{tab:AUC_compare}, we can draw the following conclusions: \textbf{(1)} The proposed LBR-SPR solution consistently outperforms state-of-the-art UVAD methods by a notable margin under all configurations. Even for the basic LBR-SPR without motion enhancement (ME), it is able to achieve $4\%$-$9\%$ AUROC gain when only videos from the testing set are used (partial mode). \textbf{(2)} Meanwhile, LBR-SPR successfully bridges the gap between UVAD and classic VAD. On these benchmarks, LBR-SPR can achieve comparable or even superior performance to latest classic VAD methods in most cases.  \textbf{(3)} Taking motion cues into consideration typically strengthens the performance of the proposed method. In particular, motion enhancement (ME) brings about $4.3\%$ AUROC gain on Avenue dataset under partial mode, which even enables LBR-SPR to yield superior performance ($92.8\%$ AUROC) to state-of-the-art classic VAD methods. \textbf{(4)} The merge mode does not necessarily produce better performance than the partial mode, e.g., on UCSDped1 and Avenue. A possible reason is that normal events in the training set are slightly different from those of the testing set. Such a distribution shift distracts DNN from reconstructing normality in the testing set, which may undermine the normality advantage and UVAD performance. 


\begin{table}
	\newcommand{\tabincell}[2]{\begin{tabular}{@{}#1@{}}#2\end{tabular}}
	\centering
	\caption{Influence of SPR on frame-level AUROC.}
	\linespread{1}\selectfont
	\scalebox{0.78}{
		\begin{tabular}{|c|c|cccc|}
			\hline
			Mode & Method & Ped1 & Ped2 & Avenue & SHTech \\
			\hline
			\multirow{2}{*}{Partial}
			& LBR  & 79.7\%  & 90.9\%  & 90.4\% & 71.7\% \\
			& LBR-SPR   & \textbf{81.1\%}  & \textbf{95.7\%}  & \textbf{92.8\%} & \textbf{72.1\%} \\
			\hline
			\multirow{2}{*}{Merge}
			& LBR  & 80.1\%  & 91.8\%  & 89.5\% & 71.7\% \\
			& LBR-SPR   & \textbf{80.9\%}   & \textbf{97.2\%}  & \textbf{90.7\%} & \textbf{72.6\%} \\
			\hline
	\end{tabular}}
	\label{tab:ablation_study}
\end{table}

\subsection{Discussion}
\label{sec:discussion}
\begin{figure}
	\centering
	\includegraphics[scale=0.75]{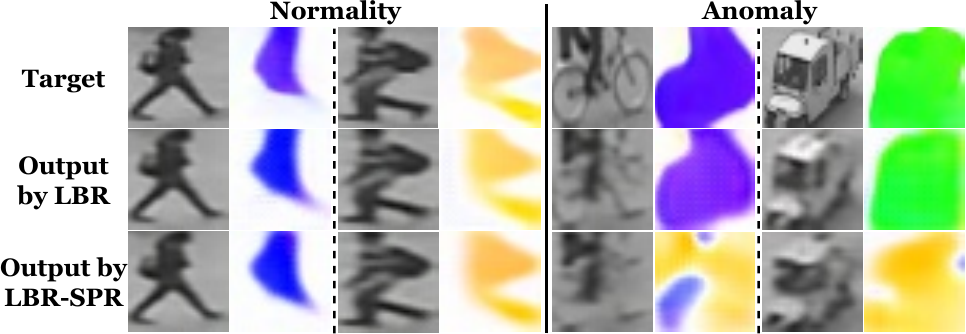}
	\caption{Reconstruction results for LBR and LBR-SPR.}
	\label{fig:visual}
\end{figure}

\textbf{Role of Self-Paced Refinement}. To demonstrate the importance of SPR to our UVAD solution, we conduct an ablation study that compares LBR and LBR-SPR under both partial and merge mode. As suggested by results in Table \ref{tab:ablation_study}, SPR constantly brings tangible performance improvement to the LBR baseline. In particular, SPR improves LBR by $4\%$ to $5\%$ AUROC on UCSDped2, as it contains a relatively high proportion of anomalies. To provide a more intuitive illustration, we further visualize some reconstruction results of LBR and LBR-SPR in Fig. \ref{fig:visual}. As shown by the figure, while LBR and LBR-SPR both reconstruct normal foreground object and its optical flow satisfactorily,  LBR-SPR reconstructs anomalies in an obviously worse manner than LBR, which makes anomalies more discriminative. 

\begin{figure}
	\centering
	\includegraphics[scale=0.61]{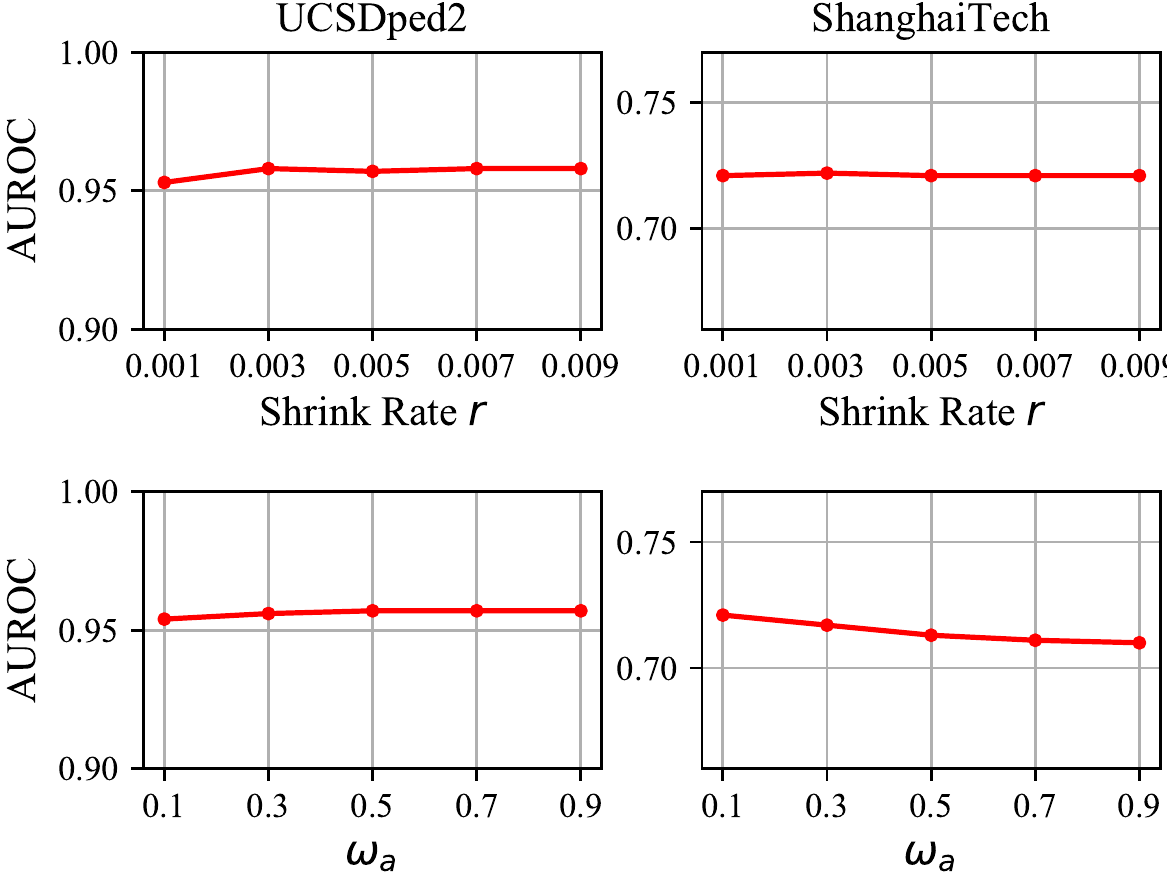}
	\caption{Parameter sensitivity analysis.}
	\label{fig:r_wa_analysis}
\end{figure}

\textbf{Sensitivity Analysis.} In Fig. \ref{fig:r_wa_analysis}, we also conduct sensitivity analysis on key parameters in our solution: \textbf{(1)} The shrink rate $r$. For demonstration, we evaluate the performance of LBR-SPR on UCSDped2 and ShanghaiTech when $r$ is varied between 0.001 and 0.01. As shown in Fig. \ref{fig:r_wa_analysis}, variation of $r$ produces up to $0.5\%$ AUROC fluctuation, which shows that the performance is not sensitive to $r$.  \textbf{(2)} Weights of anomaly scores $(\omega_a, \omega_m)$. To facilitate analysis, we simply fix $\omega_m=1$ and vary $\omega_a$ between $0.1$ and $1$ in our experiments: On UCSDped2, LBR-SPR enjoys a stable performance, while the AUROC drops by at most $1.1\%$ on ShanghaiTech when $\omega_a$ increases. However, it is noted that LBR-SPR  without ME still yields satisfactory performance.


\begin{figure}
	\centering
	\includegraphics[scale=0.59]{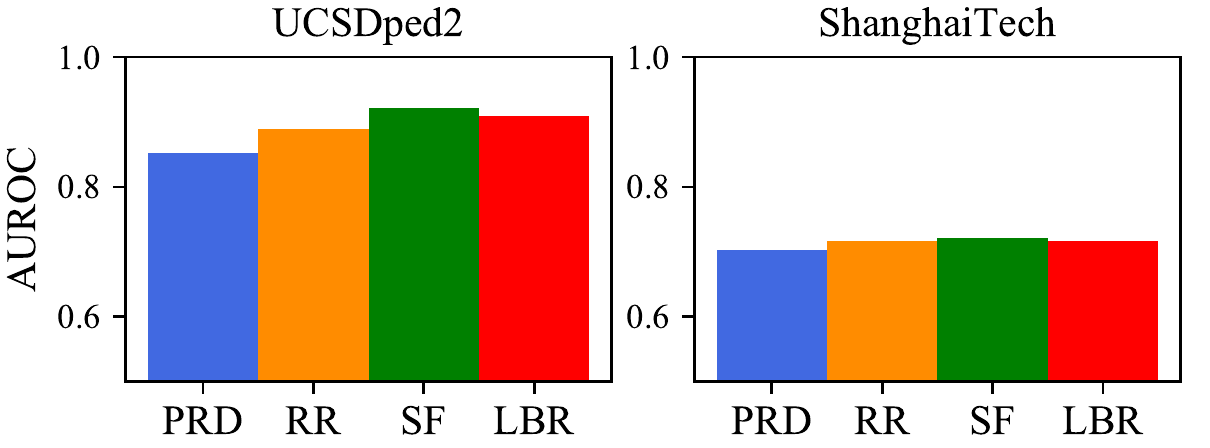}
	\caption{Other learning paradigms for UVAD.}
	\label{fig:various_tasks}
\end{figure}

\textbf{Other Learning Paradigms.} As we discussed in Sec. \ref{sec:na}, normality advantage should also be observed in other learning paradigms. To verify this, we test three additional paradigms: Prediction (PRD), reverse reconstruction (RR) and shuffling (SF). PRD aims to predict the final patch of a STC/OFC by the remaining patches; RR aims to reconstruct a STC/OFC from its reversed patch sequence; SF aims to recover a STC/OFC with randomly shuffled patches. We compare the performance of PRD/RR/SF with raw LBR on UCSDped2 and ShanghaiTech. As shown in Fig. \ref{fig:various_tasks}, other paradigms also yield close or better AUROC, which unveils the possibility to explore diverse paradigms for UVAD. More discussion are presented in supplementary material.

\section{Conclusion}

In this paper, we first reveal the advantageous role of normality in DNN based reconstruction, which enables us to propose LBR as a strong UVAD baseline. Based on LBR, we design a novel SPR scheme to remove anomalies actively, while motion cues are also exploited to further boost our solution. Our deep solution not only outperforms previous UVAD methods by a large margin, but also bridges the performance gap between UVAD and classic VAD. 

\textbf{Acknowledgements.} The work is supported by National Natural Science Foundation of China (62006236), NUDT Research Project (ZK20-10), HPCL Autonomous Project (202101-15).


{\small
\bibliographystyle{ieee_fullname}
\bibliography{egbib}
}

\end{document}